\documentclass{article}

    \usepackage[preprint]{neurips_2023}

\usepackage[utf8]{inputenc} % allow utf-8 input
\usepackage[T1]{fontenc}    % use 8-bit T1 fonts
\usepackage{hyperref}       % hyperlinks
\usepackage{url}            % simple URL typesetting
\usepackage{booktabs}       % professional-quality tables
\usepackage{amsfonts}       % blackboard math symbols
\usepackage{nicefrac}       % compact symbols for 1/2, etc.
\usepackage{microtype}      % microtypography
\usepackage{xcolor}         % colors
\usepackage{subcaption}
\usepackage{amsmath}
\usepackage{graphicx}
\title{Applications of ML-Based Surrogates in Bayesian Approaches to Inverse Problems}

\author{%
  Pelin Er\c{s}in* \\
  Izmir University of Economics \\ \\ \\
  \And 
  Emma Hayes* \\
  Carnegie Melon University \\ \\ \\
  \And
  Peter Matthews* \\
  University of Warwick \\ \\ \\
  \And 
  Paramjyoti Mohapatra* \\
  Case Western Reserve University \\ \\ \\
  \And
  Elisa Negrini \\
  University of California, Los Angeles \\ \\ \\ 
  \And
  Karl Schulz \\
  Advanced Micro Devices, Inc.
}
\newcommand\blfootnote[1]{%
  \begingroup
  \renewcommand\thefootnote{}\footnote{#1}%
  \addtocounter{footnote}{-1}%
  \endgroup
}

\begin{document}
\blfootnote{*These authors contributed equally}

\maketitle

\begin{abstract}
  Neural networks have become a powerful tool as surrogate models to provide numerical
  solutions for scientific problems with increased computational efficiency. This efficiency can
  be advantageous for numerically challenging problems where time to solution is 
  important or when evaluation of many similar analysis scenarios is required.
  One particular area of scientific interest is the setting of inverse problems, 
  where one knows the forward dynamics of a system are described by a partial differential
  equation and the task is to infer properties of the system given (potentially noisy) observations
  of these dynamics.
  We consider the inverse problem of inferring the location of a wave source on a square domain, given a
  noisy solution to the 2-D acoustic wave equation. Under the assumption of Gaussian noise,
  a likelihood function for source location can be formulated, which requires one forward simulation
  of the system per evaluation. Using a standard neural network as a surrogate model makes it 
  computationally feasible to evaluate this likelihood several times, and so Markov Chain Monte Carlo methods
  can be used to evaluate the posterior distribution of the source location. We demonstrate that this method
  can accurately infer source-locations from noisy data. 
\end{abstract}

\section{Introduction}
In Machine Learning, developing neural networks beyond the classification tasks, particularly in scientific computing, is a notable area of interest. Specifically, in recent years interest has been growing in developing neural network-based methods to solve Partial Differential Equations (PDEs) \citep{Renac2015,blechschmidt2021ways}. At present, classical numerical algorithms such as Discontinuous Galerkin methods produce results with high accuracy \citep{GALERKIN}. However, these algorithms are computationally expensive, have long run times, and often produce results on a pre-determined grid. After they have been trained, neural networks are a valuable option since they are fast 
to evaluate and can produce results on the full data domain. This efficiency can come in handy in a variety of applications, such as the inverse problem considered in this work. Previous work has used neural networks as surrogates for solving the 2D acoustic wave equation \citep{Rasht_Behesht_2022, amd2021}, but did not address the inverse problem. An interesting direction of research, which we leave for future work, is to quantify the uncertainty in these methods as done for example in \citep{Yang_2021, amd2022}

\subsection{Inverse problems}
Inverse problems involve estimating the parameters that govern a system of partial differential equations, using noisy observations of the solution, and/or forwards process, of these equations. Some work, described in \citep{dashti2015bayesian, Latz_2020}, has looked at the theory of applying inverse problems in the Bayesian setting, and others have considered Markov Chain Monte Carlo (MCMC) and surrogate modelling techniques for parameter inference \citep{Zhang2020, Zhou_2020, Wang_et_al_2020}.  A common assumption of these methods is that observations of the forwards process have Gaussian noise, which we also assume here. For the problem specified in the following subsection, using Neural Networks to model the forward process can provide the computational efficiency to tackle the inverse problem in a Bayesian setting as described in \ref{sec:inference}.

\subsection{Problem Specification}\label{section:prob}
We consider the solution to the 2-D acoustic wave equation modeled by the following PDE:
    \begin{equation}
        \frac{\partial p}{\partial t} + \kappa \cdot \nabla \boldsymbol{v} = f_s(\boldsymbol{x},t), \qquad \frac{\partial \boldsymbol{v}}{\partial t} + \frac{1}{\rho}\nabla p = 0 
    \end{equation}
    \begin{equation}
    f_s(\boldsymbol{x},t) = \frac{\tau}{\pi}\left(1 - (2\pi \omega)^2(t-t_0)^2 \right) \exp \left(-\frac{1}{2} \left[(2\pi \omega)^2 (t- t_0)^2 + 2\tau ||\boldsymbol{x} - \boldsymbol{y}_s||^2  \right] \right)
    \label{eq:Ricker}
\end{equation}

Where $\boldsymbol{x} \in [0,1] \times [0,1]$, $\boldsymbol{y_s}  \in [0,1] \times [0,1]$, and $t \in [0,2]$.
    
The forward problem is defined as predicting the pressure ($\boldsymbol{p}$) and velocity ($\boldsymbol{v}$) of the underlying wave, over the specified spatial domain over time under the influence of a forcing term ($f_s(x,t)$) . The inverse problem can now be defined as inferring the source location of the forcing term ($\boldsymbol{y_s}$), given noisy observations of pressure (or velocity) at specific receiver points. Thus, the problem at hand was broken up into two parts -- i) Designing a surrogate model that can reliably and efficiently predict $\boldsymbol{p}$ and $\boldsymbol{v}$ given $\boldsymbol{x},\boldsymbol{y_s}, t$. ii) Using the efficiency of the surrogate model to tackle the inverse problem.\\
We place a number $n$ of receivers $\left\{r_i \right\}$ in a fixed position $\boldsymbol{x}_i$ which record noisy observations of pressure $P$ at a fixed time intervals $\left\{t_j \right\}$, given possible source locations $Y_s$. We have
\begin{equation}
    P_{r_i, t_j} = p(t_j, \boldsymbol{x}_i | Y_s) + \epsilon_{i,j}, \ \ \epsilon_{i,j} \sim \mathcal{N}(0, \sigma^2_0)
\end{equation}
Where $\sigma^2_0$ is assumed to be known. Furthermore, $p(t_j, \boldsymbol{x}_i | Y_s)$ does not admit an analytical solution and we wish to use a neural-network $\tilde p(t_j, \boldsymbol{x}_i | Y_s)$ as a surrogate. This introduces further uncertainty in the model which we model as an additional independent Gaussian noise with $\sigma_1^2$. As 
to be estimated on when training the surrogate model.
\begin{equation}
    P_{r_i, t_j} = \tilde p(t_j, \boldsymbol{x}_i | Y_s) + \eta_{i,j}, \ \ \eta_{i,j} \sim \mathcal{N}(0, \sigma_0^2 + \sigma_1^2)
\end{equation}
As the support of $Y_s$ is compact, a uniform prior is well specified. We can then define the posterior $\pi(*|P)$ of source location 
given the observations, up to a normalising constant
\begin{equation}
\label{posterior}
    \pi(\boldsymbol{y}_s | P) \propto \prod_{i,j}\varphi_{\sigma_0^2 + \sigma_1^2}\left(P_{r_i, t_j} - p(t_j, \boldsymbol{x}_i | \boldsymbol{y}_s) \right)
\end{equation}
Where $\varphi_{\sigma_0^2 + \sigma_1^2}$ is the probability density of a centered 
normal distribution with variance $\sigma_0^2 + \sigma_1^2$.
\section{Methodology}
\subsection{Generating synthetic data}
The Discontinuous Galerkin Method is used to simulate the solution to the PDE from sparsely distributed source locations on the specified domain. This data it then used as ground truth data for the neural network training. The main idea is to generate ground truth data on a coarse mesh (16x16) with 50 sources distributed sparsely over the domain and then use the trained neural network to generate solutions on densely distributed source locations. 
To generate ground truth data, we use an implementation of the Discontinuous Galerkin Method similar to that of \citep{cockburndg}.We parallelized running multiple invocations of the code to generate data from multiple sources efficiently. Specifically, the simulation is run for a total of 100 different source locations on a 16x16 grid with 50 time steps. Of these 100 source simulations, 50 are used to train the neural network model and 50 are used to test the neural network's ability to predict solutions on unseen sources on the entire domain. Finally, to conduct testing on the inverse problem, Gaussian ($\sigma^2 = 0.25$) noise is added to each point in the test dataset.

\subsection{Neural Network Surrogate Model}\label{net}
\begin{figure}[h!]
\centering

    \begin{subfigure}[hb]{0.25\textwidth}
        \centering
        \includegraphics[width = \textwidth]{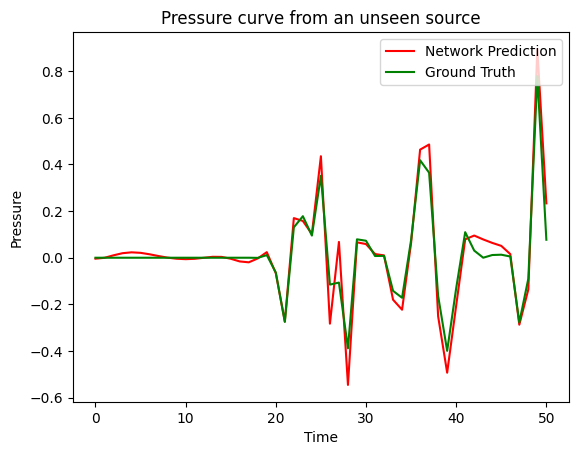}
        \caption{Pressure Trace - Unseen Source}
        \label{fig:Unseen1}
    \end{subfigure}
    \hfill
    \begin{subfigure}[hb]{0.25\textwidth}
        \centering
        \includegraphics[width = \textwidth]{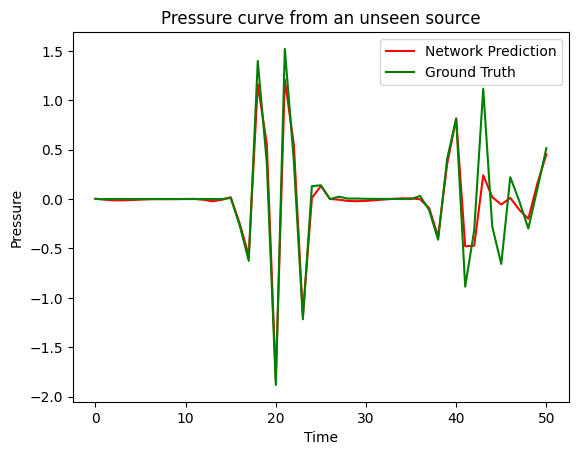}
        \caption{Pressure Trace - Unseen Source}
        \label{fig:Unseen2}
    \end{subfigure}
    \hfill
    \begin{subfigure}[hb]{0.25\textwidth}
        \centering
        \includegraphics[width = \textwidth]{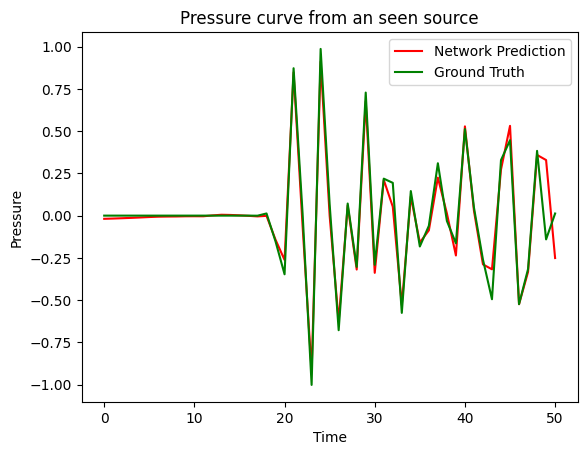}
        \caption{Pressure Trace - Seen Source}
        \label{fig:Seen1}
    \end{subfigure}
    \hfill
 
    \caption{Neural network predictions (in red) and ground truth (in green) on training and test data.}
    \label{fig:NNResults}
\end{figure}

As explained above, the first step of out algorithm is to train a Neural Network $NN(y_s, x, t, \theta)$ surrogate model to approximate the the non-linear differential operator as a solution to the set PDEs described in \ref{section:prob}. In particular, after extensive experiments with various settings and hyperparameter tuning, we selected a feed-forward network with 6 layers, each with 100 neurons, learning rate of $10^{-3}$ with decaying scheduler, batch size of 100, 2 skip connections, and tanh activation function. Figure \ref{fig:NNResults} demonstrates the network's ability to predict the solution on train and test source point.
Finally, the MSE of the network predictions on test data is used as an estimate of the variance of the surrogate noise ($\sigma_1^2 = 0.248$) for the inverse problem formulation.

\subsection{Inference of Source Position}\label{sec:inference}

The second aspect of our problem is to infer the position of the source from the solution given by the neural network surrogate model. We do this by finding a globally optimal fit to the data, and then sampling from the posterior. We first split the the domain into a $150 \times 150$ grid and evaluate the log-posterior, given sampled test data, at each point, we chose the best point on this grid as the initial point for MCMC.
 Samples from the posterior for source location were drawn using the Metropolis-Hastings algorithm \citep{Hastings_1970}. A centered multivariate Gaussian proposal distribution, with Variance-Covariance $\Sigma = 0.001 I_2$ was used and the sampler was run for 50,000 iterations. Importantly, the log-posterior is much higher at the global optimum and so we do not need to worry about exploring other modes as they have negligible probability mass. This justifies the proposed strategy of global optimisation before local sampling. We also note that the number of forward simulations required ($\approx 10^6$) makes this approach only feasible when using surrogate models like the neural network from Section \ref{net}.

\subsection{Receivers}

Due to symmetries in the spatial domain of interest, data from
from at least two receivers, that are not co-linear along an axis of symmetry, 
is needed for the source location to be identifiable. Moreover, an equal coverage of the whole spatial domain is required to reduce any systematic bias. The following set of receivers satisfy these requirements, and so they are used for numerical
experiments.
$\left\{(0.25, 0.625), (0.5, 0.5), (0.25, 0.125), (0.75, 0.625), (0.75, 0.25)\right\}.$
 
\section{Results and Discussion}
\begin{figure}[hbt]
\centering
    \begin{subfigure}[hb]{0.4\textwidth}
        \centering
        \includegraphics[width = \textwidth]{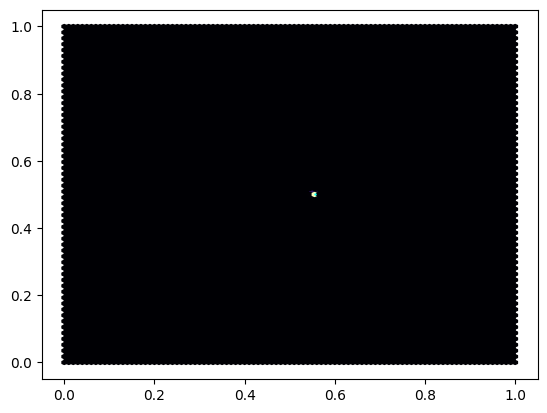}
        \caption{$Y_s = (0.556, 0.5)$, heatmap of whole domain}
        \label{fig:good_out}
    \end{subfigure}
    \hfill
    \begin{subfigure}[hb]{0.4\textwidth}
        \centering
        \includegraphics[width = \textwidth]{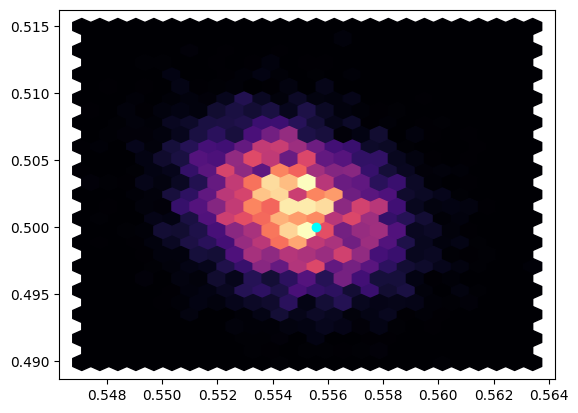}
        \caption{$Y_s = (0.556, 0.5)$, around true location}
        \label{fig:good_in}
    \end{subfigure}
    
    \begin{subfigure}[hb]{0.4\textwidth}
        \centering
        \includegraphics[width = \textwidth]{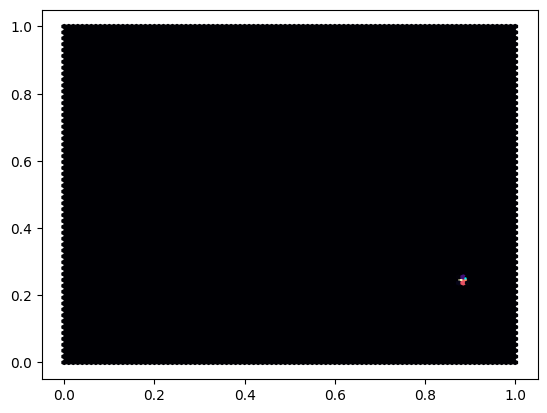}
        \caption{$Y_s = (0.889, 0.25)$, heatmap of whole domain}
        \label{fig:bad_out}
    \end{subfigure}
    \hfill
    \begin{subfigure}[hb]{0.4\textwidth}
        \centering
        \includegraphics[width = \textwidth]{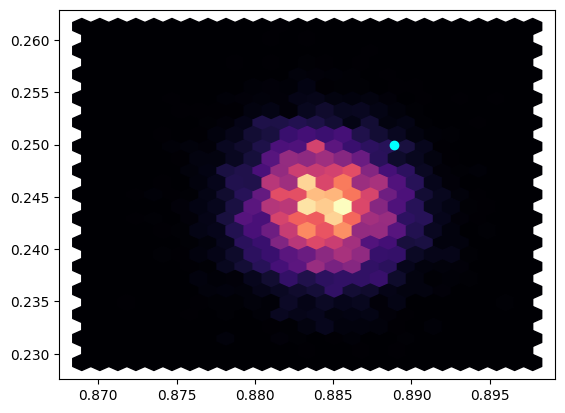}
        \caption{$Y_s = (0.889, 0.25)$, around true location}
        \label{fig:bad_in}
    \end{subfigure}
    \caption{Heatmap of samples from posterior. True locations shown in cyan.}
    \label{fig:posterior}
\end{figure}

We are able to sample from the posterior using a Metropolis Hastings algorithm that would not be feasible without neural network surrogates.
As an illustrative example, heatmaps of two different sampled posteriors are shown in Figure \ref{fig:posterior}, with true source location shown in cyan.
The posterior shows a small variance and a posterior mean very close to the true location, relative to the entire domain. However, Figure \ref{fig:bad_in} reveals common pathology. In fact, for many sources, the true location is found at the edge of the credible estimated interval, suggesting that the posterior may be overly-confident. This is due to the added uncertainty, and potentially systematic bias, introduced by using the surrogate model. This can be mitigated by increasing the estimate of $\hat{\sigma_1}^2$, however more principled ways of accounting for this uncertainty would be preferable and will be object of future work.
We also evaluate posterior samples using the mean squared error $MSE$ from the sampled source location to the true source. Across all test sources, the average relative mean squared error is $3.76 \times 10^{-5}$.

\subsection{Variation of number of Sources}
Of interest is how the accuracy and confidence of our results change with the number of receivers that the model has access to. As noted before, we require at least two receivers for an identifiable model.\\
For the true source location $(0.889, 0.25)$, Table 1 shows the variation of the MSE when decreasing the number of receivers.
As expected the prediction accuracy improves when more receivers are available, especially when adding up to 4 receivers. Adding more than 4 receivers does not seem to improve the results. One possible explanation is that, for more than 4 receivers, the dominant form of the uncertainty is due to the surrogate model error, not to the observation noise.
\vspace{-0.5cm}
\begin{table}[ht]
    \centering
    \caption{Posterior evaluations after sequentially removing receivers for true source location true source location $(0.889, 0.25)$}
    \resizebox{0.5\columnwidth}{!}{%
    \begin{tabular}{r c c} \toprule
        No. Receivers &  Posterior Mean & MSE \\ \midrule
         5 & (0.885, 0.244) & $7.94 \times 10^{-5}$ \\
         4 & (0.884, 0.246) & $7.88 \times 10^{-5}$ \\ 
         3 & (0.886, 0.243) & $1.26 \times 10^{-4}$ \\ 
         2 & (0.890, 0.241) & $1.37 \times 10^{-4}$ \\ \bottomrule
    \end{tabular}}
    \label{tab:no.receivers}
\end{table}

\section{Conclusion}
In this work we propose a technique to use neural networks as surrogates for Bayesian inverse problems. Our experiments show that, not only these surrogates are fast enough to make MCMC sampling feasible but they are also accurate enough to identify source location with a very small mean squared error (less than $10^{-4}$). Our results suggest a potential use for neural-network surrogates in solving inverse problems, whilst identifying some potential issues such as over-precise posteriors which will be addressed in a future work. Finally, the applicability of this method could further be extended by considering more complex spacial geometries or more complex samplers.

\section{Aknowledgements}
This work was conducted during the Research in Industrial Projects for Students (RIPS) program, hosted at the University of California, Los Angeles' Institute for Pure and Applied Mathematics,
from June 19 to August 18, 2023, and under the mentorship and guidance of Advanced Micro Devices, Inc. E.N. is supported by the Simons Postdoctoral program at IPAM and NSF DMS 1925919 and by AFOSR MURI FA9550-21-1-0084.
\newpage
\medskip

\bibliography{refs}
%%%%%%%%%%%%%%%%%%%%%%%%%%%%%%%%%%%%%%%%%%%%%%%%%%%%%%%%%%%%

\end{document}